\documentclass[letterpaper]{ecai}
\pdfoutput=1  

\usepackage{graphicx}
\usepackage{latexsym}

\paperid{1677}        

\usepackage{amssymb}

\usepackage{dsfont}

\usepackage[acronym]{glossaries}
\newacronym{svd}{SVD}{\emph{Singular Value Decomposition}}
\newacronym{pca}{PCA}{\emph{Principal Component Analysis}}
\newacronym{madm}{MADM}{\emph{Multi-Attribute Decision Making}}
\newacronym{iid}{i.i.d.}{independent and identically distributed}

\newacronym{ml}{PL}{Pseudo-Likelihood}
\newacronym{ev}{EV}{Embedded Voting}
\newacronym{ml+}{PL+}{Pseudo-Likelihood +}
\newacronym{ev+}{EV+}{Embedded Voting +}
\newacronym{av}{AV}{Approval Voting}
\newacronym{rv}{RV}{Range Voting}
\newacronym{np}{NP}{Nash Product}
\newacronym{ga}{MA}{Model-Aware}
\newacronym{sa}{SA}{Single Agent}
\newacronym{rw}{RW}{Random Winner}

\newacronym{normal}{N}{Normal}
\newacronym{wide}{W}{Wide}
\newacronym{thin}{T}{Thin}
\newacronym{left}{L}{Left Skewed}
\newacronym{right}{R}{Right Skewed}

\glsdisablehyper

\newcommand{\BibTeX}{\rm B\kern-.05em{\sc i\kern-.025em b}\kern-.08em\TeX}

\newcommand{\A}{\mathcal{A}} 
\newcommand{\C}{\mathcal{C}} 
\newcommand{\R}{\mathbb{R}} 
\newcommand{\E}{\mathbb{E}} 
\newcommand{\N}{\mathcal{N}} 
\newcommand{\Prob}{\mathbb{P}} 

\newcommand{\estEmbeddingVector}[1]{\hat{\vec{e_{#1}}}} 
\newcommand{\estEmbeddingVectorIJ}{\hat{e}_{i,j}} 
\newcommand{\estScoreMatrix}[1]{\hat{M}_{#1}} 
\newcommand{\estK}{\hat{k}} 

\usepackage{pgfplots}
\pgfplotsset{compat=1.17,
	legend image code/.code={
		\draw[mark repeat=2,mark phase=2]
		plot coordinates {
			(0cm,0cm)
			(0.15cm,0cm)        
			(0.3cm,0cm)         
		};
	},
	curve/.style = {thick, mark=*, mark size=1, mark options={solid}},
	mylegend/.style = {font=\tiny,
		fill opacity=1,
		draw opacity=1,
		text opacity=1,
		draw=white!80!black,
	},
	fivecolumns/.style = {   legend columns=2, transpose legend},
	twocolumns/.style = {	legend columns=5,
		transpose legend,
	},
	sixentries/.style = {	legend columns=3,
		transpose legend,
	}
}
\usepackage[subrefformat=parens,labelformat=parens]{subfig}
\usepackage{stfloats}
\usepgfplotslibrary{groupplots}

\usepackage{xcolor}
\definecolor{nasht}{rgb}{0.658823529411765,0.23921568627451,0.23921568627451}
\definecolor{mlet}{rgb}{0.270588235294118,0.250980392156863,0.811764705882353}
\definecolor{nash}{rgb}{0.870588235294118,0.188235294117647,0.164705882352941}
\definecolor{app}{rgb}{0.196078431372549,0.901960784313726,0.172549019607843}
\definecolor{sum}{rgb}{0.870588235294118,0.87843137254902,0.274509803921569}
\definecolor{mle}{rgb}{0.141176470588235,0.533333333333333,0.929411764705882}
\colorlet{model}{white!43.921568627451!black}
\colorlet{single}{white!55!black}
\colorlet{random}{white!65!black}

\usepackage{hyperref}

\begin{document}

\begin{frontmatter}

\title{Aggregating Correlated Estimations\\with (Almost) no Training}

\author[A]{\fnms{Théo}~\snm{Delemazure}\thanks{Corresponding Author. Email: \href{mailto:theo.delemazure@dauphine.eu}{theo.delemazure@dauphine.eu}}
}
\author[B]{\fnms{François}~\snm{Durand}}
 \author[C]{\fnms{Fabien}~\snm{Mathieu}} 

\address[A]{LAMSADE, Paris Dauphine University, PSL, CNRS}

\address[B]{Nokia Bell Labs France}
\address[C]{Swapcard}

\begin{abstract}
Many choice problems cannot be solved exactly and use several estimation algorithms that assign scores to the different available options. 
The estimation errors can have various correlations, from low (e.g. between two very different approaches) to high (e.g. when using a given algorithm with different hyperparameters). Most aggregation rules would suffer from this diversity of correlations.
In this article, we introduce \emph{\gls{ev}}, an aggregation rule designed to take correlations into account, and we compare it to other aggregation rules in various experiments based on synthetic data. Our results show that when sufficient information about the correlations between errors is available, a maximum likelihood aggregation should be preferred. Otherwise, typically with limited training data, \gls{ev} outperforms the other approaches.
\end{abstract}

\end{frontmatter}

\vspace{-.2cm}
\section{Introduction}
\label{sec:intro}
\vspace{-.1cm}

We consider the problem of score aggregation through the general choice problem of having to choose one item from a set of options called \emph{candidates}. The actual utility of each candidate is unknown and impossible to compute, but we have access to human or software agents that can provide a noisy estimate of it. How to select the best possible candidate?

In the aggregated output, the errors of the individual agents should ideally mitigate one another and produce a collective decision that would outperform the one of a single agent, like in the seminal Condorcet Jury Theorem~\cite{condorcet1785essai}. However, two issues can hamper aggregation. The first one is the varying accuracy of the agents. 
Although of great importance, this issue is outside the scope of this paper, where we focus on the second one: correlations between agents. 

This problem can arise from various reasons. In the case of software agents, it is frequent to use several variants of an algorithm, or the same algorithm with different hyperparameters. In the case of human experts, some of them may share a common sociocultural background or prior deliberation on the topic at hand, which may give them a common bias.

These correlations, which may be unknown to the aggregation rule, may hinder its performance. Consider the following toy example: Three algorithms are used, two of them being exact clones (they always return the same estimate). Unaware of the clones, the aggregation rule uses the median score. As the median estimate is always the one of a clone, the aggregation behaves exactly as any of the clones, and the potential benefit from the independence of the third algorithm is lost.

\vspace{-.2cm}
\subsection{Related Work}\label{sec:related-work}
\vspace{-.2cm}

Our model is close to the one of \emph{epistemic social choice}, in which voters have noisy estimates of the candidates' true values. Some works investigate the quality of the results of specific voting rules in this model~\cite{procaccia1,caragiannis1,conitzer1}.  

Decisions based on the aggregation of scores appear in many contexts, not necessarily related to the human activity of \emph{social choice}. In regression analysis, the quality of prediction can be improved by aggregating the results of different algorithms. For example, the methods of \emph{boosting} and \emph{bagging} achieve some form of aggregation~\cite{sutton2005classification}.

When several estimations are available for the same problem, it is common for some of them to be correlated, for example when the same technique is used with different hyperparameters or different training datasets. Since these correlations can be harmful for aggregation, 
many of the main methods require the panel of algorithms to be \emph{diversified}~\cite{GRANITTO2005139}. In particular, this avoids biasing the results in the direction of a potential \emph{``majority group''}. Here, we do not assume this diversity constraint to be satisfied.

Our topic is also close to the Unsupervised Ensemble Learning problem, where the goal is to find the best way to aggregate the outputs of a family of classifiers without any feedback. Some methods in this area address the problem of correlated classifiers \cite{unsupervisedEnsembleLearning1,unsupervisedEnsembleLearning2,kleindessner2018crowdsourcing}. These methods are also useful in the crowd-sourcing framework in which agents might be correlated.
However, in their setting, each choice problem always examines the same candidates (e.g. ``cat'' and ``dog'') in different situations (e.g. images to classify), whereas in our setting, there is no obvious mapping between the candidates of any two distinct choice problems.

Finally, there is a vast literature in statistics applied to physics on the problem of aggregating correlated measures of the same quantity to reduce the noise 
\cite{bayescombiningcorrelatedestimates,HELENE201682,averagingcorrelateddata,hiddencorrelations}. In particular, the method called BLUE is equivalent to the model-aware benchmark that we present in Section~\ref{sec:likelihood} \cite{combinationcorrelatedphysics,Nisius_2020}. But our general problem differs as we do not assume any prior knowledge on the distributions and the correlations of the estimators. Moreover, we focus on the choice problem of selecting the best candidate, which is not the same as inferring one value from noisy estimates. We also differentiate ourselves by proposing a novel method based on spectral decomposition for this problem.

It is worth noting that this paper introduces a new choice rule where the identification of correlations between agents is based on the classic data preprocessing technique known as \gls{svd} \cite{Wall2003}. The use of \gls{svd} is not uncommon in related fields such as Multi-armed Bandits \cite{auer2003linrel} (to compute upper confidence bounds) and recommendation systems~\cite{svdcf} (for collaborative filtering). However, to the best of our knowledge, this is the first time that \gls{svd} is applied to a choice problem based on noisy estimations by correlated agents.

\vspace{-.1cm}
\subsection{Contribution}
\vspace{-.1cm}

In this paper, we propose a new aggregation technique, \acrfull{ev}, inspired by the notion of \emph{Nash product} (cf.~Section~\ref{sec:nash}). \gls{ev} uses spectral analysis 
to mitigate the impact of correlations. We validate the interest of this technique in a controlled model where agents only differ by the correlations between them, but not by their accuracy. We compare it to other approaches, using as upper bound a maximum likelihood technique based on a full knowledge of the model parameters.

Our main findings are that with a prior training, \gls{ev} is on par with the upper bound, except when all noise parameters are high (cf. Section~\ref{sec:noise-intensity}); without training, \gls{ev} is the best of all the untrained aggregation rules of our benchmark throughout all our experiments, with a performance that is generally very close to its trained version, meaning that it can generally infer the relevant information about the correlations just by considering the options of the present choice problem.

The rest of the paper is organized as follows.
Section~\ref{sec:score-aggregation} presents our proposal for score aggregation. 
Section~\ref{sec:validation} gives a first experimental validation. 
Section~\ref{sec:impact-of-parameters} explores how various parameters impact the performance of the proposed solutions. 
Section~\ref{sec:conclusion} concludes.

\section{Score Aggregation}\label{sec:score-aggregation}

Our notations are inspired by the field of \emph{computational social choice}~\cite{brandt2016handbook}. We consider a choice problem with a given set $\C = \{c_1, \ldots, c_m\}$ of $m$ candidates. Each candidate $c_j \in \C$ has a utility $u(c_j) \in \R$. We want to select a candidate with the highest possible utility.

However, the utility values cannot be directly observed
. Instead, we have access to a set $\A = \{a_1, \ldots, a_n \}$ of $n$ agents that provide utility estimates. Formally, each agent $a_i$ assigns a 
\emph{score} to each candidate $c_j$, denoted by
$s_i(c_j) \in \R
$. 
We also introduce $\varepsilon_i(c_j) := s_i(c_j) - u(c_j)$, the (unknown) noise of agent $a_i$ for candidate $c_j$. We view $\varepsilon_i(c_j)$ as a random variable to enable the use of a probabilistic framework.
All the scores can be represented in an $n\times m$ matrix $S=(s_i(c_j))_{1\le i\le n, 1\le j\le m}$. An \emph{aggregation rule} is a function $f$ that associates to $S$ a candidate $f(S)\in\C$. 
Our goal is to maximize $\E\big(u(f(S))\big)$.
The rest of this section presents different approaches to solve this problem.

\vspace{-.1cm}
\subsection{Welfare-Based Approaches}\label{sec:classical-approaches}
\vspace{-.1cm}

In many contexts, the \emph{utilitarian welfare} is a good metric to select a winner from a list of candidates. 
In our case, viewing the scores as subjective utility values, the corresponding aggregation rule is as follows: compute the welfare of each candidate as the sum of its scores, $w_{\text{util}}(c_j) := \sum_{i=1}^n s_i(c_j)$; then choose a candidate with maximal welfare.
This rule is equivalent to selecting the candidate with the highest average score. It is usually called \emph{\gls{rv}} in the literature of computational social choice \cite{smith2000range}.

If the noise functions $\varepsilon_i$ are drawn from independent and identical Gaussian distributions, then \gls{rv} is the optimal aggregation rule (see Theorem 4.1 of~\cite{kunchevabook} or Section~\ref{sec:likelihood}). 
However, 
we precisely focus on cases where they are not independent. For example, consider $n=24$ agents such that $20$ of them are strongly correlated and each of the remaining 4 is independent of all the others. In most cases, the utilitarian welfare will mostly depend on the scores given by the large group of 20 agents, and the information given by the $4$ independent agents will mostly be ignored. 
More generally, the lack of diversity can significantly decrease the quality of the results given by the utilitarian welfare maximization~\cite{GRANITTO2005139,diversity}. 

A variant of \gls{rv} consists in processing the raw scores before summing them in order to mitigate the impact of large errors. For example, we can apply a Heaviside step function shifted by a given threshold
. In that case, the welfare of a candidate is simply the number of agents that rate the candidate above the threshold, which is a form of \emph{\gls{av}} \cite[p.~53]{brandt2016handbook}. Note that \gls{av} increases the odds of having multiple candidates that maximize the welfare, especially if the number $n$ of agents is small. Therefore, the choice of the tie-breaking rule may be important.

Another notion of welfare is the \emph{\gls{np}}, widely used in game theory~\cite{kaneko1979nash,NASH1}. It has appealing fairness properties in numerous contexts, like collective decision making or project fundings~\cite{aziz2019fair,BRANDL2022102585,caragiannis2019unreasonable,fluschnik2019fair}. Assuming that all scores are positive
, the Nash product of a candidate is defined as the product of its scores: $w_{\text{Nash}}(c_j) := \prod_{i=1}^n s_i(c_j)$. 
It also suffers in the case of varying correlations: for example, if one independent agent mistakenly doubles the score of a candidate, then the welfare of the candidate is doubled, but if a group of $k$ correlated agents do the same, then the welfare of the candidate is multiplied by~$2^k$. 

\vspace{-.1cm}
\subsection{Maximum-Likelihood Approaches}\label{sec:ml-approaches}
\vspace{-.1cm}

Viewing the noises as random variables, it is natural to use a maximum likelihood estimator~\cite{eliason1993maximum}. With perfect information about the joint distribution of the noises, it can give the best possible estimates for the utility values of the candidates (Section~\ref{sec:likelihood} gives an example of exact computation). However, it is impossible in general to determine the characteristics of these distributions from the observed data alone. For example, consider the following case with $n=m=2$: $s_1(c_1)=s_2(c_1)=10$ and $s_1(c_2)=s_2(c_2)=15$, i.e. both agents estimate that the first candidate has utility $10$ and the second has utility~$15$. It is possible that both agents are error-free and that $10$ and $15$ are the two utility values. But it is also possible that both agents are noisy and fully correlated, that both candidate have the same utility, and that the difference between the values $10$ and $15$ just comes from the shared error of the agents. The two cases are impossible to distinguish from observation alone. 

\vspace{-.1cm}
\subsection{Our Proposal: Embedded Voting}
\label{sec:nash}
\vspace{-.1cm}

We now introduce our aggregation rule, whose goal is to exploit correlations between the agents without requiring prior knowledge on the ground truth or the noise model.

To start with, consider the following ideal case: the set of agents $\A$ is partitioned into $k$ perfectly identified groups $A_1, \dots, A_k$. The agents of the same group are fully correlated (they always yield the same estimates) while agents from different groups are independent. All the agents have the same accuracy. We focus on the extraction of some welfare value $w(c_j)$ associated to a given fixed candidate $c_j$, based on the estimates $s_i(c_j)$ as in Section~\ref{sec:classical-approaches}. 

Contrary to most classical voting rules, which are \emph{anonymous}, in the sense that they treat voters symmetrically~\cite{brandt2016handbook}, here we want to use the additional information that we have about the agents in order to weigh them. In particular a desired property would be that the impact of a given group of identical agents does not depend on its size. A natural way to achieve this is to assign the same weight to each group. If using the utilitarian welfare, this leads to the average of the average scores of each group, which results in the following welfare for candidate~$c_j$: 
\vspace{-.2cm}\begin{align}\label{eq:perfect_mean}
w(c_j) := \frac{1}{k} \sum_{l=1}^{k} \frac{1}{\left | A_l \right |} \sum_{a_i \in A_l} s_i(c_j)\text{.}
\end{align}

If the scores are positive
, one can also consider the Nash product of the utilitarian welfare of each group, defined as:
\vspace{-.2cm}\begin{align}\label{eq:perfect_groups}
w(c_j) := \prod_{l=1}^{k} \sum_{a_i \in A_l} s_i(c_j)\text{.}
\end{align}

Equation~\ref{eq:perfect_groups} is affected by the size of the groups, like Equation~\ref{eq:perfect_mean}, but with the same scaling factor for each candidate: for example, if we double the size of one group, it simply doubles the welfare of each candidate, which does not change the outcome of the choice problem.

However, in practice, groups cannot be precisely delimited: the correlations within the groups may be hard to identify, or some agents may share correlations with several groups. Yet, it is possible to use spectral analysis to infer the underlying groups and generalize Equation~\ref{eq:perfect_groups} to arbitrary cases.

But keep on considering the ideal case for the moment. We can represent the link between agents and groups by vectors: to each agent $a_i \in A_l$, associate the $1\times k$ vector $\vec{e_i} := (\mathbf{1}_{l'=l})_{1\leq l' \leq k}$ ($1$ at the $l^\text{th}$ position, $0$ everywhere else). We call $\vec{e_i}$ the \emph{embedding} of agent~$a_i$.

Using the embeddings, define the $n\times k$ score matrix $M_j$ of candidate $c_j$ as follows: 
\begin{eqnarray}
	\label{eq:Mj}
	M_j := 
	\begin{pmatrix}
		\sqrt{s_{1}(c_j)}\vec{e_1}  \\
		\vdots \\
		\sqrt{s_{n}(c_j)}\vec{e_n}  
	\end{pmatrix}\text{.}
\end{eqnarray}

Denote by $(\lambda_1, \ldots, \lambda_k)$ the singular values of the matrix $M_j$, obtained by \gls{svd}~\cite{Wall2003}. 
Up to re-indexing, 
we have the relation ${\lambda_l}^2 = \sum_{i: a_i \in A_l} s_i(c_j) $ (the square root in \eqref{eq:Mj} provides the benefit of this simple formula). Thus, each ${\lambda_l}^2$ represents the utilitarian welfare of the group $A_l$. In particular, the product $\prod_{l=1}^k{\lambda_l}^2$ is exactly the welfare value defined by Equation~\ref{eq:perfect_groups}.

Using spectral analysis may seem excessive in the ideal case where the partition is perfect and known. However, its main advantage is that it is based on some embedding of the agents that can convey more information than a simple partition. In particular, we can extend Equation~\ref{eq:perfect_groups} to a more general case if we can infer an embedding of the agents.

To this end, we simply define the estimated embedding vectors $\estEmbeddingVector{i}$ of the agents as their score vectors (including the scores on a potential training set of candidates), normalized such that all the score vectors of the agents have the same mean and standard deviation. More specifically, we have the $j^\text{th}$ component of agent $i$ set as:
\begin{eqnarray}
    \label{eq:eij}
    \estEmbeddingVectorIJ{} := \frac{s_i(c_j) - \text{avg}_{j'} s_i(c_{j'})}{\text{std}_{j'} s_i(c_{j'})}.
\end{eqnarray}
Then for each candidate $c_j$, we define $\estScoreMatrix{j}$ similarly to Equation~\eqref{eq:Mj}. Note that its size is $n \times m$, instead of $n \times k$ for $M_j$ in the ideal case.

For a given candidate $c_j$, denote $(\lambda_1, \ldots, \lambda_{\min(m, n)})$ the singular values of $\estScoreMatrix{j}$ sorted in descending order.
$\estScoreMatrix{j}$ can contain redundant dimensions (in particular if the agents are correlated), so we would like to compute the welfare $w(c_j)$ as the product $\prod_{l=1}^{\estK} {\lambda_l}^2$ of the $\estK$ greatest singular values, where $\estK$ would estimate the number of relevant dimensions in the embedding of the agents.

To estimate $\estK$, we define $\hat{S} = (\estEmbeddingVectorIJ)_{i, j}$, the normalized full score matrix, and we set $\estK$ to the number of singular values above $0.95$ times the average of all $\min(m, n)$ singular values of~$\hat{S}$.\footnote{
Ideally, we would like to only keep the singular values that are above the average of the singular values. But if all the voters are independent, we would like to keep all the singular values, i.e. $\estK=n$. However, in that case, all the singular values will be very similar but not perfectly equal due to noise and numerical errors, so some of them will be below the average, and thus we would have $\estK<n$. The 0.95 discount enables these singular values that are slightly below the average to still be taken into account. Thus, we will get the correct $\estK$ in the symmetric case without impacting much the non-symmetric cases.
}

We call the resulting method \emph{\acrfull{ev}}. 

As an illustration, let us apply this method to the ideal case, in the limit of a large dataset (i.e. large $m$). By definition in Equation~\ref{eq:eij}, all vectors $\estEmbeddingVector{i}$ have norm 1. When two agents are fully correlated, they have the same embedding vector. And when two agents $i$ and $i'$ are independent, then by definition of independence, their vectors $\estEmbeddingVector{i}$ and $\estEmbeddingVector{i'}$ are orthogonal in the limit of a large dataset. So the estimated embedding $(\estEmbeddingVector{i})_i$ will be the same as $(\vec{e_i})_i$, up to an orthonormal change of basis. It is then easy to prove that the singular values of $\hat{S}$ tend to $(|A_1|, \ldots, |A_k|, 0, \ldots, 0)$ in the limit of a large dataset. The average singular value is $(\sum_{l} |A_l|) / n = 1$, so the number $\estK$ of singular values higher than 0.95 the average is precisely $k$. Now consider a candidate~$j$. Once again, $\estScoreMatrix{j}$ is the same as $M_j$, up to an orthonormal change of basis and addition of irrelevant dimensions due to the statistical noise. So taking the $\estK = k$ greatest singular values of $\estScoreMatrix{j}$ will give the same result as relying on $M_j$.

\section{Validation}\label{sec:validation}

We now evaluate the performance of the aggregation rules that we have described. We use artificial datasets to have a full control over the behavior of the candidates and the agents. Broadly speaking, one performance measure goes as follows:
\begin{itemize}
	\item Draw for $m$ candidates their utilities $u(c_j), 1\leq j \leq m$.
	\item Use a noise model to draw the agent estimations $\big(s_i(c_j))_{1\leq i \leq n, 1\leq j \leq m}$.
	\item Feed different aggregation rules with the estimates (the rules never access the utility values).
	\item Measure with the utility values the quality of the returned candidate.
	\item Repeat the steps above enough times to have a good measurement of the expected performance.
\end{itemize}

In details, we draw the utility values of the candidates independently according to a normal distribution, which is chosen to be $\mathcal{N}(0, 1)$ without loss of generality. By default, for each choice problem, $m=20$ candidates are considered.

The performance metric we use to compare the different aggregation rules is the \emph{average relative utility}: for each choice, if a rule yields candidate $c_j$, then its relative utility is a value between 0 and 1 defined by  $\frac{u(c_j) - u_{\min}}{u_{\max} - u_{\min}}$, where $u_{\max}$ (resp. $u_{\min}$) is the best (resp. worst) utility among those of the candidates. 
For each rule, we compute the average relative utility over 10,000 choices to assert its performance.\footnote{In our preliminary experiments, we also compared the different rules with respect to the accuracy metric, i.e. the probability to choose the best candidate, and we observed similar trends than for utility and relative utility.} The open source code developed for this paper
is available on GitHub\footnote{\url{https://github.com/TheoDlmz/embedded_voting}.} .

The rest of this section is organized as follows. Section~\ref{sec:profile-model} details our noise model. Section~\ref{sec:likelihood} provides the exact maximum-likelihood solution, which requires the knowledge of the model and its parameters, along with an approximation, which does not. Section~\ref{sec:rules} recaps the aggregation rules that we consider and gives some implementation details. Section~\ref{sec:study-of-a-base-scenario} presents a first representative experiment.

\subsection{Noise Model}\label{sec:profile-model}

We want our noise model to have the following properties:
\begin{itemize}
	\item All agents have the same accuracy (we focus here on correlations). Considering agents with different accuracies and weighting them accordingly is another complex problem, outside the scope of this paper.
	\item The correlations between agents can be tuned finely.
\end{itemize}

To achieve these objectives, we propose the following model, which combines correlated and uncorrelated errors. It relies on three parameters:
\begin{itemize}
	\item An $n\times k$ matrix $E=(e_{i, l})_{ 1\leq i \leq n, 1\leq l \leq k}$
for some $k \in \mathbb N$ that represents the agents in a \emph{feature} space.
	\item A \emph{feature} noise intensity $\sigma_f \in \R_{\ge 0}$.
	\item A \emph{distinct} noise intensity $\sigma_d \in \R_{\ge 0}$.
\end{itemize}

The matrix $E$ generalizes the embedding introduced in Section~\ref{sec:nash}. For convenience, we assume that each row of $E$ has at least one non-zero component and that its Euclidean norm is 1: $\forall i \in \{1, \ldots, n\}, \sum_{1\leq l \leq k}{e_{i, l}}^2=1$. The parameter $\sigma_f$ determines the part of the noise related to the underlying features of the agents, while $\sigma_d$ determines some noise that is independent across all agents.

For a given candidate $c_j$, we draw the noises as follows: we first draw $n+k$ i.i.d. values $d_{1,j}, \ldots, d_{n,j}, f_{1,j}, \ldots, f_{k, j}$ following the standard normal distribution $\N(0, 1)$. The first $n$ values determine noises that are specific to each agent, while the last $k$ ones are specific to each feature. The estimation noise of agent $a_i$ for $c_j$ is then defined as:
\begin{equation}
	\varepsilon_i(c_j):=\sigma_d d_{i,j} + \sigma_f\sum_{1\leq l \leq k}e_{i, l}f_{l, j}\text{.}
\end{equation} 

By construction, the marginal distribution of each $\varepsilon_i(c_j)$ is the sum of two independent distributions $\N(0, \sigma_d)$ and $\N(0, \sigma_f)$, and two given agents have fully independent noises if they have no feature in common. 

Equivalently, let $S(c_j)$ be the $n\times 1$ random vector of the scores provided by the agents, $\bar{E}:=\begin{pmatrix}
	\sigma_d I_n & \sigma_f E
\end{pmatrix}$ be the $n\times (n+k)$ matrix that concatenates the characteristics of distinct noise and feature noise (where $I_n$ denote the $n \times n$ identity matrix), and $X_j=(d_{1, j},\ldots, d_{n, j}, f_{1, j}, \ldots, f_{k, j})^T$ be a random vector whose $n + k$ values follow i.i.d. standard normal distributions. Under our noise model, $S(c_j)$ can be written as:
\begin{equation}\label{eq:distrib_s_cj}
	S(c_j) = u(c_j)\mathds{1}_{n \times 1} + \bar{E} X_j\text{.}
\end{equation}
Thus, conditionally on the value of $u(c_j)$, $S(c_j)$ follows a multivariate normal distribution of mean $u(c_j)\mathds{1}_{n \times 1}$ and covariance matrix $\Sigma:= \bar{E}\bar{E}^T$.

Let us illustrate the noise model on a few toy examples: in the degenerate case where $E$ is a $n \times 0$ matrix, i.e. $\bar{E} = I_n$, we have $n$ independent agents; the case $E=\begin{pmatrix}
	1 & 0 \\
	1 & 0 \\
	0  & 1
\end{pmatrix}$ with $\sigma_d = 0$ models the example with two clones given in the introduction (more generally, the ideal case with $k$ perfect groups introduced in Section \ref{sec:nash} can be modeled with a $n\times k$ indicator matrix); the case $E=\begin{pmatrix}
	1 & 0 \\
	0 & 1 \\
	1/\sqrt{2} &  1/\sqrt{2}
\end{pmatrix}$ models two independent agents and a third one that is correlated with the other two.

\subsection{Maximum Likelihood}\label{sec:likelihood}

With the knowledge of the noise model, i.e. $(E, \sigma_d, \sigma_f)$, it is possible to infer the maximum likelihood utility of a candidate $c_j$ given the observed estimates. For the moment, let us ignore our knowledge on the distribution of utility values, and assume that $u(c_j)$ is a hidden parameter of the model. Hence the only probabilistic contribution comes from the noise.

Denoting $s_j:=(s_1(c_j), \ldots, s_n(c_j))^T$ a score vector and $\Delta(s_j, u(c_j))=\left(s_j - u(c_j)\mathds{1}_{n \times 1}\right)$, Equation~\ref{eq:distrib_s_cj} implies:
\begin{equation}
	\Prob\big(S(c_j)=s_j\big) \propto e^{-\frac{1}{2}\Delta(s_j, u(c_j))^T\,\Sigma^{-1}\,\Delta(s_j, u(c_j))}\text{.}
\end{equation}

To find the maximum likelihood estimation $u_j^*$ of the unknown value $u(c_j)$, we just need to maximize the above expression. 
Denoting $\omega=(\omega_1, \ldots, \omega_n)$ the vector $\mathds{1}_{1 \times n}\Sigma^{-1}$, logarithmic derivation gives the following solution:
\begin{equation}\label{eq:likelihood}
	u_j^* = \frac{\sum_{i=1}^n \omega_i s_i(c_j)}{\sum_{i=1}^n \omega_i}\text{.}
\end{equation}
Thus, $u_j^*$ is just a well-chosen weighted average of the observed scores. Intuitively, the weights are lower for correlated agents to account for their redundancy. One can check that in the case of fully correlated groups (each agent belongs to a single feature, no distinct noise), the weight of an agent is the inverse of the size of its group. In particular, we have the intuitive 
result that if all agents are independent (each belong to a group of size 1), the simple average of the observed score is the best estimator.

Now assume that we also know the distribution of the utility values.
For simplicity, we also assume that its probability density function (PDF) can be written as $\exp(f(u))$, where $f$ is twice differentiable. 
By a similar likelihood maximization computation as above, we find a new solution $u^+_j$ that meets:
\begin{equation}\label{eq:likelihood_knowing_distrib_truth}
    0 = f'(u_j^+) + \Omega(u_j^*-u_j^+)\text{, with }  \Omega := \sum_{i=1}^n \omega_i.
\end{equation}
Implicitly, this defines $u_j^+$ as a function of $u_j^*$. By implicit differentiation, we obtain:
\begin{equation}
    \frac{\mathrm{d} u_j^*}{\mathrm{d} u_j^+} = 1- \frac{f''(u_j^+)}{\Omega}.
\end{equation}
If $f''$ is always smaller than $\Omega$ (and in particular if it is nonpositive), then $u_j^*$ is increasing wrt~$u_j^+$. This means that selecting the best candidate according to $u^+$ or according to $u^*$ is equivalent. In other words, the selection of the best candidate is not improved by knowing the utility distribution, as long as its PDF is log-concave, which is the case for a normal distribution.

We call \acrfull{ga} the rule that selects the candidate with (equivalently) the highest $u_j^+$ or $u_j^*$
. Since it relies on the best possible estimate (in the sense of maximum likelihood), we argue that \gls{ga} should be a near-perfect solution. In our particular use case, it happens to be equivalent to the method BLUE \cite{combinationcorrelatedphysics,Nisius_2020}, even if the rationale behind BLUE is based on minimizing the variance and not on maximizing likelihood.

That being said, our paper assumes that the noise model is unknown to the observer in practice. Hence, while \gls{ga} gives a nice tool to estimate an upper bound of performance, it is not a usable rule as it cannot be computed only from the observed scores. 

However, the covariance matrix $\Sigma$ of the noise can be loosely approximated from the covariance $\hat{\Sigma}$ of the observed scores, which combines the noise of the agents and the utility variations of the candidates. The difference between $\Sigma$ and $\hat{\Sigma}$ comes from two factors: $\Sigma$ and $\hat{\Sigma}$ do not correspond to the same random variable; $\hat{\Sigma}$ is an estimate based on observation.

In details, the utility variations introduce a global correlation between agents, so the random variable measured by $\hat{\Sigma}$ is not the true noise: all correlations are over-estimated. If the noise dominates utility variation, it stays an accurate approximation. On the other hand, if the noise is small compared to utility variation, the over-estimation of agent correlations will make the weights $\omega$ in Equation~\ref{eq:likelihood} sub-optimal (they will be more uniform than if $\Sigma$ was used), but we argue that if the noise is small the impact on performance should be limited.

Also note that if the number of candidates is low, $\hat{\Sigma}$ is noisy due to the lack of observations.
Using $\hat{\Sigma}$ instead of $\Sigma$ gives a rule that can be used in practice, which we call \gls{ml}.

\subsection{Summary of the Rules Considered}\label{sec:rules}

We can now summarize all the rules that we want to evaluate, and specify some implementation details.

Unless otherwise stated, each agent rescales its scores to have a null mean and a unit standard deviation, so that agents with different score scales can be compared. This is not necessary in our experimental setting where all agents have the same marginal distribution but it may be required in practical applications.

All the rules that we study aggregate the estimated scores in a welfare measure $w$ in order to select the winning candidate.

\sloppy

For	\acrfull{rv}, the welfare is $w(c_j) := \sum_{a_i\in \A}s_i(c_j)$.

For \acrfull{av}, the welfare is $w(c_j) := \sum_{a_i\in \A}\mathbf{1}_{s_i(c_j)\ge 0}$: each agent computes a threshold as its average score across candidates and approve of all the candidates with a higher score. \gls{av} often generates ties, so we investigated several reasonable tie-breaking rules, like the sum, the product of the positive scores, or a fixed arbitrary order over the candidates. The results, not displayed in this paper, are relatively similar with a slight advantage for the product of positive scores, which we will then use.

The \acrfull{np} and \acrfull{ev} require non-negative estimates. To do so, for both rules, each agent first rescales its observations to have unit standard deviation and mean 2 (i.e. two standard deviations).

For \gls{np}, we need positive values so the estimates that are still below $0.1$ are replaced with $0.1$. Then $w(c_j) := \prod_{a_i \in \mathcal{A}} s_i(c_j)$.

For \gls{ev}, we only need non-negative values so the threshold is set to 0 instead. \gls{ev} then uses $w(c_j) := \prod_{l=1}^k{\lambda_l}^2$, where $\lambda_1,\ldots,\lambda_k$ are the $\estK$ greatest singular values of the score matrix $\estScoreMatrix{j}$ and $\estK$ is the estimated number of relevant dimensions, computed as described in Section~\ref{sec:nash}. The embedding vectors $(\estEmbeddingVector{i})_i$ of the agents, used to define $\hat{S}$ and $\estScoreMatrix{j}$, are based on the matrix of scores of the agents on the candidates considered for the problem. If we have access to other candidates, like candidates considered in previous choices, we can increase the precision of the method by considering all of them (from past and current decisions) to build the embedding vector $\estEmbeddingVector{i}$ of each agent~$i$.
This defines a \emph{trained} version of \gls{ev} based on previous observations, which we call \gls{ev+}.

\acrfull{ga} computes the welfare from Equation~\ref{eq:likelihood}, where the weight vector $\omega$ is deduced from the noise model. As the underlying noise model cannot be known by a rule in general, \gls{ga} is essentially an upper reference point. We also consider \acrfull{ml}, which approximates the noise model from the observed scores. As for \gls{ev}, it is possible to use past candidates to enhance the precision of the approximation. We call \gls{ml+} the \emph{trained} version of \gls{ml} based on previous observations, 
which uses all candidates from previous and current decisions to build the estimate $\hat{\Sigma}$ of the covariance matrix.

Unless otherwise stated, \gls{ev+} and \gls{ml+} use a training set of 1,000 candidates (not including the 20 candidates of the current choice).

Additionally, we consider \gls{sa}, a rule that uses the estimated score of a single agent (no gain from aggregation) and \gls{rw}, a rule that selects a candidate randomly regardless of the estimations. These two rules serve as lower reference points for the other rules. 

\subsection{Study of a Reference Scenario}\label{sec:study-of-a-base-scenario}

We start our evaluation with the same setting as in the example of Section~\ref{sec:classical-approaches}: a correlated group of $20$ agents and $4$ additional independent agents. This case can be described with a $24 \times 5$ embedding matrix
defined by blocks as  $E = \left( \begin{array}{cc} \mathds{1}_{20\times 1} & 0 \\ 0 & I_4 \end{array} \right)$, where $I_4$ is the $4 \times 4$ identity matrix.
We use a feature noise intensity $\sigma_f = 1$ (i.e. equal to the standard deviation of utility values between candidates) and a distinct noise intensity $\sigma_d = 0.1$. The avowed purpose of this setting is to demonstrate the problems caused by correlations, by considering a large correlated group and a high feature noise compared to the distinct noise. 

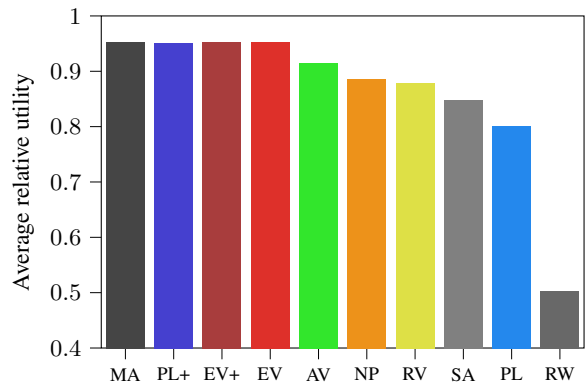
\begin{figure}[!h]
\begin{tikzpicture}

\definecolor{color0}{rgb}{0.270588235294118,0.250980392156863,0.811764705882353}
\definecolor{color1}{rgb}{0.658823529411765,0.23921568627451,0.23921568627451}
\definecolor{color2}{rgb}{0.870588235294118,0.188235294117647,0.164705882352941}
\definecolor{color3}{rgb}{0.196078431372549,0.901960784313726,0.172549019607843}
\definecolor{color4}{rgb}{0.929411764705882,0.572549019607843,0.101960784313725}
\definecolor{color5}{rgb}{0.870588235294118,0.87843137254902,0.274509803921569}
\definecolor{color6}{rgb}{0.141176470588235,0.533333333333333,0.929411764705882}

\begin{axis}[
height=6cm, width=8cm,
tick align=outside,
tick pos=left,
x grid style={white!69.0196078431373!black},
xmin=-0.5, xmax=9.5,
xtick style={color=black},
xtick={0,1,2,3,4,5,6,7,8,9},
xticklabel style={font=\scriptsize},
xticklabels={
  \gls{ga},
  \gls{ml+},
  \gls{ev+},
  \gls{ev},
  \gls{av},
  \gls{np},
  \gls{rv},
  \gls{sa},
  \gls{ml},
  \gls{rw}
},
y grid style={white!69.0196078431373!black},
ylabel={Average relative utility},
ymin=0.4, ymax=1.,
ytick style={color=black},
ytick distance=0.1
]
\draw[draw=none,fill=white!27.0588235294118!black] (axis cs:-0.4,0) rectangle (axis cs:0.4,0.953234546628265);
\draw[draw=none,fill=color0] (axis cs:0.6,0) rectangle (axis cs:1.4,0.950327950359675);
\draw[draw=none,fill=color1] (axis cs:1.6,0) rectangle (axis cs:2.4,0.952188919876275);
\draw[draw=none,fill=color2] (axis cs:2.6,0) rectangle (axis cs:3.4,0.951934546119433);
\draw[draw=none,fill=color3] (axis cs:3.6,0) rectangle (axis cs:4.4,0.915632571535544);
\draw[draw=none,fill=color4] (axis cs:4.6,0) rectangle (axis cs:5.4,0.886747995562417);
\draw[draw=none,fill=color5] (axis cs:5.6,0) rectangle (axis cs:6.4,0.878504737197228);
\draw[draw=none,fill=white!50.1960784313725!black] (axis cs:6.6,0) rectangle (axis cs:7.4,0.848397516218121);
\draw[draw=none,fill=color6] (axis cs:7.6,0) rectangle (axis cs:8.4,0.800640278285634);
\draw[draw=none,fill=white!40.7843137254902!black] (axis cs:8.6,0) rectangle (axis cs:9.4,0.502488018548346);
\end{axis}

\end{tikzpicture}

	\caption{Performance in our reference scenario.\label{fig:base}}
\end{figure}

The average relative utility for each rule is displayed in Figure~\ref{fig:base}. The lower reference point is \gls{rw} with 50\%. Clearly, as long as the distribution of utility values is symmetric, it will always be the case, and we will omit \gls{rw} from the figures. As expected, the upper reference point is \gls{ga}, with 95\%.

We first observe the relatively poor performance of \gls{rv} (88\%), which is only slightly above the use of a single agent (85\%). This is a case of \emph{tyranny of the majority} \cite{tocqueville1840democratie}: 
the group of $20$ agents concentrates most of the decision power but its \emph{collective wisdom} is close to the one of a single agent because of the correlations. \gls{np} has similar performance (89\%). 
Surprisingly, the approval variant of \gls{rv}, \gls{av}, performs significantly better despite its simplicity (92\%). Intuitively, as \gls{av} compresses the range of estimations, it prevents a large error from the group in favor of a candidate to dominate the decision.

Then, we see that the three rules \gls{ev}, \gls{ev+}, and \gls{ml+} have remarkable performance (95\%). The gain compared to the next best rule, \gls{av}, is small in percentage (3\%), but important as we are close to the optimal: in terms of \emph{regret}, i.e. difference with the true optimal choice, we go from 8\% to only 5\%. Moreover, the difference with the upper reference \gls{ga} is within the margin of error (which is of the order of 1\%, because the sample size is 10,000). 

On the other hand, the performance of \gls{ml} is remarkably poor (80\%): still much better than returning a random candidate (50\%) but worse that only relying on the estimate of one single agent (85\%). 

\section{Impact of the Parameters}\label{sec:impact-of-parameters}

In order to examine more broadly the performance of the rules considered, we now alter some parameters in isolation, while the others keep their values from the reference scenario. 

\subsection{Number of Agents}\label{sec:number-of-experts}

\begin{figure}[!h]
\begin{tikzpicture}

\definecolor{color0}{rgb}{0.270588235294118,0.250980392156863,0.811764705882353}
\definecolor{color1}{rgb}{0.658823529411765,0.23921568627451,0.23921568627451}
\definecolor{color2}{rgb}{0.870588235294118,0.188235294117647,0.164705882352941}
\definecolor{color3}{rgb}{0.196078431372549,0.901960784313726,0.172549019607843}
\definecolor{color4}{rgb}{0.929411764705882,0.572549019607843,0.101960784313725}
\definecolor{color5}{rgb}{0.870588235294118,0.87843137254902,0.274509803921569}
\definecolor{color6}{rgb}{0.141176470588235,0.533333333333333,0.929411764705882}

\begin{axis}[
height=6cm, width=8cm,
legend cell align={left},
legend style={mylegend, twocolumns,
  at={(0.03,0.03)},
  anchor=south west,
},
tick align=outside,
tick pos=left,
width=8cm,
x grid style={white!69.0196078431373!black},
xlabel={Size of the correlated group},
x label style = {yshift=.15cm},
xmajorgrids,
xmin=1, xmax=30,
xtick style={color=black},
y grid style={white!69.0196078431373!black},
ylabel={Average relative utility},
ymajorgrids,
ymin=0.686472242781277, ymax=0.96644878242217,
ytick style={color=black}
]
\addplot [thick, white!27.0588235294118!black, mark=*, mark size=1, mark options={solid}]
table {
1 0.953542187605252
2 0.953722576074857
4 0.953250768887607
6 0.953293243046743
8 0.953301805190407
10 0.95325056100751
12 0.95324835266766
14 0.953330279651625
16 0.953232321685405
18 0.953269815363474
20 0.953234546628264
22 0.95324457401582
24 0.953258485760457
26 0.953195866197356
28 0.95330229296089
30 0.953304525725978
};
\addlegendentry{\gls{ga}}
\addplot [thick, color0, mark=*, mark size=1, mark options={solid}]
table {
1 0.952636163723342
2 0.952420820057007
4 0.952597655137868
6 0.952037697504606
8 0.9518162377416
10 0.951474984539261
12 0.951420325935643
14 0.951037784945935
16 0.950938090414413
18 0.950584276330134
20 0.950327950359674
22 0.950269313314097
24 0.949323087041885
26 0.948897057060327
28 0.948912951006336
30 0.949262108101465
};
\addlegendentry{\gls{ml+}}
\addplot [thick, color1, mark=*, mark size=1, mark options={solid}]
table {
1 0.953656764778717
2 0.952509987611379
4 0.952371260887858
6 0.952182259484695
8 0.952287580207867
10 0.952178364930857
12 0.952083512276953
14 0.952328249324306
16 0.952275917991457
18 0.952276875565374
20 0.952188919876276
22 0.952166105564007
24 0.952170180535344
26 0.952129347042171
28 0.952105655646854
30 0.952173105020766
};
\addlegendentry{\gls{ev+}}
\addplot [thick, color2, mark=*, mark size=1, mark options={solid}]
table {
1 0.946065341601677
2 0.948138782555466
4 0.95138566703919
6 0.952096406008132
8 0.951949942906544
10 0.951941858989944
12 0.952036390416615
14 0.951985306276907
16 0.951957905139339
18 0.951955290473946
20 0.951934546119435
22 0.951970313869253
24 0.951871157427357
26 0.951886731248763
28 0.951814733889316
30 0.951943936005404
};
\addlegendentry{\gls{ev}}
\addplot [thick, color3, mark=*, mark size=1, mark options={solid}]
table {
1 0.943027261368771
2 0.941378272610651
4 0.935585756338819
6 0.930199742002057
8 0.926321540717207
10 0.923856161009325
12 0.921779267953716
14 0.919601676551796
16 0.918142653086684
18 0.916413331730935
20 0.915632571535547
22 0.9148554869556
24 0.914096437862102
26 0.913504794204252
28 0.912899316529464
30 0.912320425863539
};
\addlegendentry{\gls{av}}
\addplot [thick, color4, mark=*, mark size=1, mark options={solid}]
table {
1 0.952187187322946
2 0.948597507622118
4 0.935937016038796
6 0.923684600684167
8 0.914455496263614
10 0.907156207318968
12 0.90148115602333
14 0.897493948968277
16 0.89365280039143
18 0.889865166277037
20 0.886747995562418
22 0.884443429835947
24 0.882576764067746
26 0.880549030767976
28 0.879088807884363
30 0.877410068990618
};
\addlegendentry{\gls{np}}
\addplot [thick, color5, mark=*, mark size=1, mark options={solid}]
table {
1 0.952670182606584
2 0.94850888649262
4 0.931732802016204
6 0.916224356457924
8 0.90544626906459
10 0.898298114323499
12 0.892207153788652
14 0.887914749247664
16 0.884069388557924
18 0.880794394189487
20 0.87850473719723
22 0.876455145051521
24 0.874449607344446
26 0.872476620210336
28 0.871009721391584
30 0.869726223356944
};
\addlegendentry{\gls{rv}}
\addplot [thick, white!50.1960784313725!black, mark=*, mark size=1, mark options={solid}]
table {
1 0.847131013027302
2 0.84952238009808
4 0.849439852193992
6 0.848503567627618
8 0.847681263994487
10 0.848290990058918
12 0.848181908919775
14 0.849269872564645
16 0.849713314252859
18 0.848550280116668
20 0.848397516218122
22 0.848845465679748
24 0.848765525876671
26 0.847925351465993
28 0.847663795447411
30 0.848309081194633
};
\addlegendentry{\gls{sa}}
\addplot [thick, color6, mark=*, mark size=1, mark options={solid}]
table {
1 0.938259302554239
2 0.931286549561397
4 0.914862259799552
6 0.894219065865398
8 0.867050860590792
10 0.831864378770482
12 0.78068800146972
14 0.704769083134938
16 0.69919844912859
18 0.763690901619437
20 0.800640278285636
22 0.827015897434366
24 0.840247650123689
26 0.854277005193464
28 0.862942557215869
30 0.870308662991849
};
\addlegendentry{\gls{ml}}
\end{axis}

\end{tikzpicture}

	\caption{Changing the size of the correlated group.\label{fig:group}}
\end{figure}
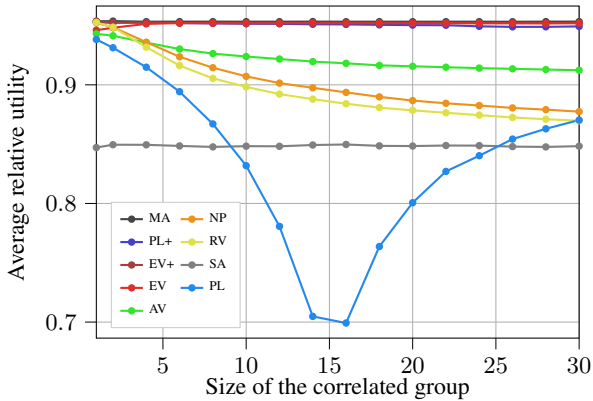

To start with, Figure~\ref{fig:group} varies the size of the correlated group, which is $20$ in the reference scenario, from~$1$ (i.e. $5$ independent agents) to $30$. Other things being equal, a larger correlated group should provide marginally better performance for optimal methods: the law of large numbers tend to cancel out the distinct noises inside the correlated group. On the other hand, a larger correlated group also pollutes the pool of scores with redundant information.

\interfootnotelinepenalty=10000

We first observe that \gls{ml+}, \gls{ev+}, and \gls{ev} are very stable at 95\%, within the margin of error with \gls{ga}: they manage to constantly ignore the redundant information. \gls{rv} is optimal in the independent case but its performance decreases for a large group (87\%), where it tends to ignore the scores of the independent agents. \gls{np} shows similar results (88\% for a large group). \gls{av} is slightly sub-optimal in the independent case (94\%), but it is more robust and becomes competitive for large groups (91\%). Lastly, \gls{ml} is insufficiently trained with the candidates of the current choice only and behaves erratically, with minimal performance around 15 correlated agents due to ranks effects in the correlation matrix\footnote{When $m \approx n$, the conditioning of the estimated correlation matrix $\hat{\Sigma}$ is bad. With noise, its pseudo-inverse creates artifacts in the weight vector $\omega$ used in Equation~\ref{eq:likelihood}: some components become negative so the result is out of the convex hull of the agents' estimations. This explains why \gls{ml} behaves worse than selecting a single method.}. 

\begin{figure}[!h]
\begin{tikzpicture}

\definecolor{color0}{rgb}{0.270588235294118,0.250980392156863,0.811764705882353}
\definecolor{color1}{rgb}{0.658823529411765,0.23921568627451,0.23921568627451}
\definecolor{color2}{rgb}{0.870588235294118,0.188235294117647,0.164705882352941}
\definecolor{color3}{rgb}{0.196078431372549,0.901960784313726,0.172549019607843}
\definecolor{color4}{rgb}{0.929411764705882,0.572549019607843,0.101960784313725}
\definecolor{color5}{rgb}{0.870588235294118,0.87843137254902,0.274509803921569}
\definecolor{color6}{rgb}{0.141176470588235,0.533333333333333,0.929411764705882}

\begin{axis}[
height=6cm, width=8cm,
legend cell align={left},
legend style={mylegend, fivecolumns,
  at={(0.97,0.03)},
  anchor=south east,
},
tick align=outside,
tick pos=left,
width=8cm,
x grid style={white!69.0196078431373!black},
xlabel={Number of independent agents},
x label style = {yshift=.15cm},
xmajorgrids,
xmin=0, xmax=20,
xtick style={color=black},
y grid style={white!69.0196078431373!black},
ylabel={Average relative utility},
ymajorgrids,
ymin=0.731664017663485, ymax=0.999290155159377,
ytick style={color=black}
]
\addplot [thick, white!27.0588235294118!black, mark=*, mark size=1, mark options={solid}]
table {
0 0.849881242811159
1 0.90292202121364
2 0.928959466592291
4 0.953234546628264
6 0.964385130068372
8 0.971983486605628
10 0.976504200627347
12 0.980207823290631
14 0.982359502972196
16 0.984616381258191
18 0.985764104718282
20 0.987125330727745
};
\addlegendentry{\gls{ga}}
\addplot [thick, color0, mark=*, mark size=1, mark options={solid}]
table {
0 0.849570929397765
1 0.901529621631006
2 0.926865453637827
4 0.950327950359674
6 0.961236250208462
8 0.968409073007138
10 0.973963423461769
12 0.977061594425903
14 0.979563391067188
16 0.981261973410096
18 0.982603445957531
20 0.98356161809758
};
\addlegendentry{\gls{ml+}}
\addplot [thick, color1, mark=*, mark size=1, mark options={solid}]
table {
0 0.849881242811159
1 0.901953703028503
2 0.927645985100915
4 0.952188919876276
6 0.96341798772402
8 0.971186649551768
10 0.975771275126379
12 0.979668200243325
14 0.981393244418688
16 0.983684153386136
18 0.985395675648682
20 0.986579575693918
};
\addlegendentry{\gls{ev+}}
\addplot [thick, color2, mark=*, mark size=1, mark options={solid}]
table {
0 0.849783829901809
1 0.890657578171804
2 0.923669049075531
4 0.951934546119435
6 0.962206337451029
8 0.969710469750536
10 0.974695907076327
12 0.978229972234237
14 0.980128697087333
16 0.981990192735614
18 0.983901803699066
20 0.985193632384105
};
\addlegendentry{\gls{ev}}
\addplot [thick, color3, mark=*, mark size=1, mark options={solid}]
table {
0 0.849900967538724
1 0.876543213967209
2 0.893825090943428
4 0.915632571535547
6 0.928566609787658
8 0.938719315896799
10 0.94553117571197
12 0.951846506691931
14 0.956012747825615
16 0.959526829166486
18 0.962833353207451
20 0.965039197291724
};
\addlegendentry{\gls{av}}
\addplot [thick, color4, mark=*, mark size=1, mark options={solid}]
table {
0 0.849875422967255
1 0.860596549958795
2 0.87010304974717
4 0.886747995562418
6 0.900281349068651
8 0.911001228774538
10 0.919808440954547
12 0.928545162674552
14 0.934792351418711
16 0.940043341676312
18 0.945119542805139
20 0.949814658900634
};
\addlegendentry{\gls{np}}
\addplot [thick, color5, mark=*, mark size=1, mark options={solid}]
table {
0 0.849884327898019
1 0.857862543704468
2 0.864479574227881
4 0.87850473719723
6 0.89010325472288
8 0.899527048542039
10 0.908473229828132
12 0.916975952521597
14 0.924145635985148
16 0.928975196507371
18 0.934907950634731
20 0.940353596051386
};
\addlegendentry{\gls{rv}}
\addplot [thick, white!50.1960784313725!black, mark=*, mark size=1, mark options={solid}]
table {
0 0.848397516218122
1 0.848397516218122
2 0.848397516218122
4 0.848397516218122
6 0.848397516218122
8 0.848397516218122
10 0.848397516218122
12 0.848397516218122
14 0.848397516218122
16 0.848397516218122
18 0.848397516218122
20 0.848397516218122
};
\addlegendentry{\gls{sa}}
\addplot [thick, color6, mark=*, mark size=1, mark options={solid}]
table {
0 0.840249936041196
1 0.743828842095117
2 0.766873968514127
4 0.800640278285636
6 0.825669709488629
8 0.838079965521522
10 0.841758627017151
12 0.836734289642814
14 0.82294890698233
16 0.810788227727587
18 0.803824072627496
20 0.819351019400994
};
\addlegendentry{\gls{ml}}
\end{axis}

\end{tikzpicture}

	\caption{Changing the number of uncorrelated agents.\label{fig:indep}}
\end{figure}
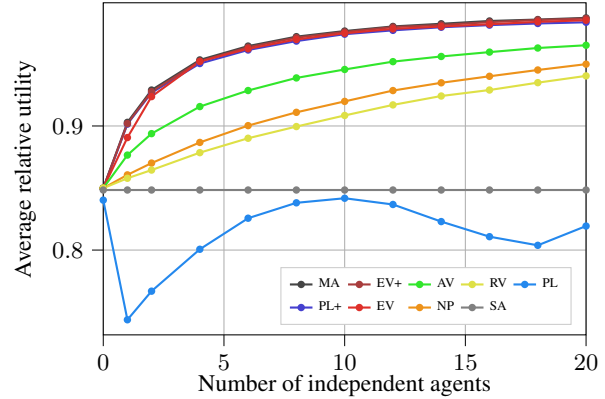

Figure~\ref{fig:indep} shows the companion experiment investigating the impact of the number of independent agents (4 in the reference scenario). More independent agents give more information. The best rules \gls{ml+}, \gls{ev+} and \gls{ev} take advantage of the additional information, and so do \gls{av}, \gls{np} and \gls{rv} to a lesser extent. \gls{ml} is constantly worse than a single agent.

\subsection{Number of Candidates}\label{sec:number-of-candidates}

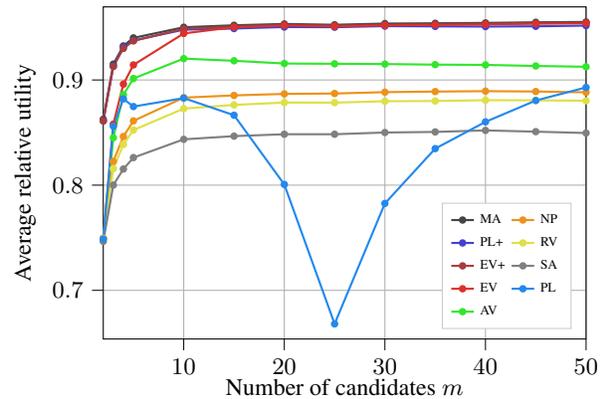
\begin{figure}[!h]
\begin{tikzpicture}

\definecolor{color0}{rgb}{0.270588235294118,0.250980392156863,0.811764705882353}
\definecolor{color1}{rgb}{0.658823529411765,0.23921568627451,0.23921568627451}
\definecolor{color2}{rgb}{0.870588235294118,0.188235294117647,0.164705882352941}
\definecolor{color3}{rgb}{0.196078431372549,0.901960784313726,0.172549019607843}
\definecolor{color4}{rgb}{0.929411764705882,0.572549019607843,0.101960784313725}
\definecolor{color5}{rgb}{0.870588235294118,0.87843137254902,0.274509803921569}
\definecolor{color6}{rgb}{0.141176470588235,0.533333333333333,0.929411764705882}

\begin{axis}[
height=6cm, width=8cm,
legend cell align={left},
legend style={mylegend, twocolumns,
  at={(0.97,0.03)},
  anchor=south east,
},
tick align=outside,
tick pos=left,
width=8cm,
x grid style={white!69.0196078431373!black},
xlabel={Number of candidates \(\displaystyle m\)},
x label style = {yshift=.15cm},
xmajorgrids,
xmin=2, xmax=50,
xtick style={color=black},
y grid style={white!69.0196078431373!black},
ylabel={Average relative utility},
ymajorgrids,
ymin=0.653741473322231, ymax=0.969477741016243,
ytick style={color=black}
]
\addplot [thick, white!27.0588235294118!black, mark=*, mark size=1, mark options={solid}]
table {
2 0.8626
3 0.915043823546875
4 0.932345415679633
5 0.939992230543564
10 0.950093850355387
15 0.951971265097648
20 0.953234546628264
25 0.952461012637129
30 0.953587587490528
35 0.953861244528
40 0.954216579397724
45 0.95480166586444
50 0.955126092484696
};
\addlegendentry{\gls{ga}}
\addplot [thick, color0, mark=*, mark size=1, mark options={solid}]
table {
2 0.8611
3 0.913118809328278
4 0.93135718390055
5 0.937547664579342
10 0.947791504177289
15 0.948893545413961
20 0.950327950359674
25 0.950147785161973
30 0.951189241072319
35 0.950930416498199
40 0.950748867398665
45 0.950999547080268
50 0.951677454015599
};
\addlegendentry{\gls{ml+}}
\addplot [thick, color1, mark=*, mark size=1, mark options={solid}]
table {
2 0.8611
3 0.912870153578907
4 0.929992983674125
5 0.937127177512376
10 0.948600610898124
15 0.950783540130011
20 0.952188919876276
25 0.951440772475066
30 0.952395006725285
35 0.95276916901216
40 0.953070481832101
45 0.953637423147796
50 0.954087504045131
};
\addlegendentry{\gls{ev+}}
\addplot [thick, color2, mark=*, mark size=1, mark options={solid}]
table {
2 0.7495
3 0.857987901305189
4 0.8960492307839
5 0.914298339925172
10 0.944225587785753
15 0.95019226395123
20 0.951934546119435
25 0.950980390063602
30 0.951938909010529
35 0.952369853317239
40 0.95288569022469
45 0.953584710925521
50 0.953985534041877
};
\addlegendentry{\gls{ev}}
\addplot [thick, color3, mark=*, mark size=1, mark options={solid}]
table {
2 0.7487
3 0.845109722056104
4 0.886018792332561
5 0.901381824154121
10 0.920293111297137
15 0.918233592698961
20 0.915632571535547
25 0.915364713757488
30 0.915114458547128
35 0.914563504148623
40 0.914327611763601
45 0.91330600150481
50 0.912529626158146
};
\addlegendentry{\gls{av}}
\addplot [thick, color4, mark=*, mark size=1, mark options={solid}]
table {
2 0.7495
3 0.822655867418984
4 0.846257655020484
5 0.86112150852316
10 0.883195307464497
15 0.885310864598307
20 0.886747995562418
25 0.88712572392067
30 0.888419809956821
35 0.888972512302302
40 0.889356766715872
45 0.889026487830106
50 0.888250255431121
};
\addlegendentry{\gls{np}}
\addplot [thick, color5, mark=*, mark size=1, mark options={solid}]
table {
2 0.7494
3 0.815669045335828
4 0.83867657988755
5 0.852424249675207
10 0.872754432824285
15 0.876302816181982
20 0.87850473719723
25 0.878445658672326
30 0.879779710341331
35 0.879994493119386
40 0.880765017076688
45 0.880666074947868
50 0.880227924990859
};
\addlegendentry{\gls{rv}}
\addplot [thick, white!50.1960784313725!black, mark=*, mark size=1, mark options={solid}]
table {
2 0.7466
3 0.800071734620045
4 0.815317060623836
5 0.826194420935399
10 0.843552372981279
15 0.846603002574872
20 0.848397516218122
25 0.848405532249803
30 0.85004572109055
35 0.850609378421993
40 0.85203909987548
45 0.850855755783174
50 0.849562189128226
};
\addlegendentry{\gls{sa}}
\addplot [thick, color6, mark=*, mark size=1, mark options={solid}]
table {
2 0.7488
3 0.856078173308889
4 0.882029347760699
5 0.87471839020237
10 0.882752721513663
15 0.866492227493919
20 0.800640278285636
25 0.668093121853777
30 0.782609113186387
35 0.834709488291685
40 0.86016584217863
45 0.880450672146176
50 0.892901389658741
};
\addlegendentry{\gls{ml}}
\end{axis}

\end{tikzpicture}

	\caption{Changing the number of candidates.\label{fig:candi}}
\end{figure}

Figure~\ref{fig:candi} varies the number of candidates. More candidates means more ``good'' ones, hence a rule can afford to choose a good candidate even if it misses the best one. This explains the global increase of relative utility with the number of candidates. In the extreme case of 2 candidates, the non-trained \gls{ev} is significantly less efficient (75\%) than \gls{ga}, \gls{ml+} and \gls{ev+} (86\%), but with at least 10 candidates, the difference is within the margin of error: \gls{ev} efficiently exploits the candidates of the current election as a training set to identify the correlations between agents. The performance of \gls{ml} is minimal around $m=n$, for the same reason as in Figure~\ref{fig:group}.

\vspace{-.1cm}
\subsection{Noise Intensity}\label{sec:noise-intensity}
\vspace{-.1cm}

\renewcommand{\thefigure}{6}
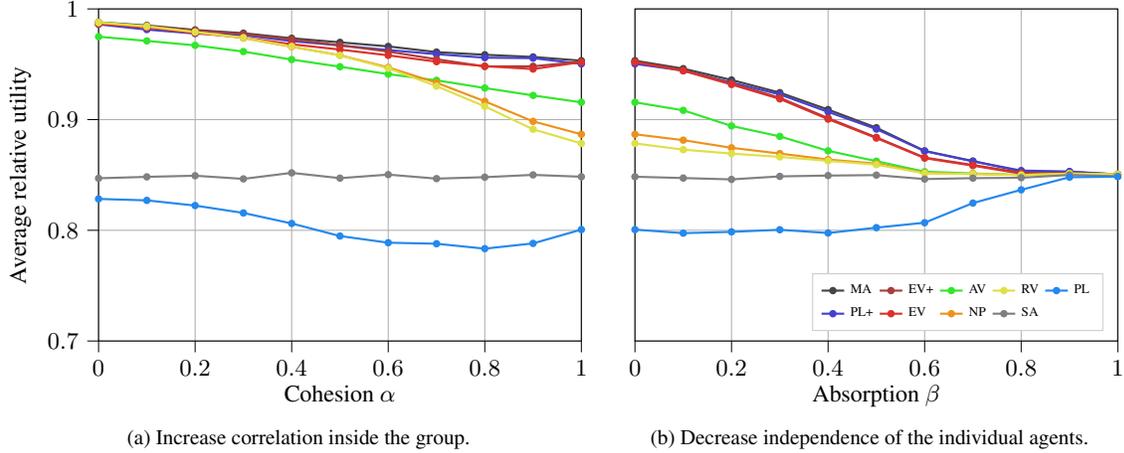
\begin{figure*}[!t]
	\centering
	\subfloat[Increase correlation inside the group.\label{fig:alpha}]{
\begin{tikzpicture}

\definecolor{color0}{rgb}{0.270588235294118,0.250980392156863,0.811764705882353}
\definecolor{color1}{rgb}{0.658823529411765,0.23921568627451,0.23921568627451}
\definecolor{color2}{rgb}{0.870588235294118,0.188235294117647,0.164705882352941}
\definecolor{color3}{rgb}{0.196078431372549,0.901960784313726,0.172549019607843}
\definecolor{color4}{rgb}{0.929411764705882,0.572549019607843,0.101960784313725}
\definecolor{color5}{rgb}{0.870588235294118,0.87843137254902,0.274509803921569}
\definecolor{color6}{rgb}{0.141176470588235,0.533333333333333,0.929411764705882}

\begin{axis}[
height=6cm, width=8cm,
tick align=outside,
tick pos=left,
width=8cm,
x grid style={white!69.0196078431373!black},
xlabel={\strut Cohesion \(\displaystyle \alpha\)},
x label style = {yshift=.15cm},
xmajorgrids,
xmin=0, xmax=1,
xtick style={color=black},
y grid style={white!69.0196078431373!black},
ylabel={Average relative utility},
ymajorgrids,
ymin=0.7, ymax=1,
ytick style={color=black}
]
\addplot [thick, white!27.0588235294118!black, mark=*, mark size=1, mark options={solid}]
table[row sep=\\] {
0 0.987949096503821 \\
0.1 0.98506954216209 \\
0.2 0.98100044504452 \\
0.3 0.978058983762306 \\
0.4 0.973491087437948 \\
0.5 0.969801914826934 \\
0.6 0.966121475485659 \\
0.7 0.960988430633855 \\
0.8 0.958575608550386 \\
0.9 0.956601672131167 \\
1 0.953234546628265 \\
};
\addlegendentry{\gls{ga}}
\addplot [thick, color0, mark=*, mark size=1, mark options={solid}]
table[row sep=\\] {
0 0.986061833606353 \\
0.1 0.981340892657318 \\
0.2 0.977748457563413 \\
0.3 0.976282827418733 \\
0.4 0.970901957740772 \\
0.5 0.966978216231637 \\
0.6 0.96286031538499 \\
0.7 0.959111408762535 \\
0.8 0.955958388195646 \\
0.9 0.955502257913892 \\
1 0.950327950359675 \\
};
\addlegendentry{\gls{ml+}}
\addplot [thick, color1, mark=*, mark size=1, mark options={solid}]
table[row sep=\\] {
0 0.986452411834885 \\
0.1 0.984348763156017 \\
0.2 0.980510484690633 \\
0.3 0.977198588586447 \\
0.4 0.972520708042419 \\
0.5 0.967150661252026 \\
0.6 0.961465961167884 \\
0.7 0.95449416600632 \\
0.8 0.948023818993724 \\
0.9 0.948160534077458 \\
1 0.952188919876275 \\
};
\addlegendentry{\gls{ev+}}
\addplot [thick, color2, mark=*, mark size=1, mark options={solid}]
table[row sep=\\] {
0 0.986859505163506 \\
0.1 0.983224944605467 \\
0.2 0.978242607391887 \\
0.3 0.97416377995568 \\
0.4 0.968072920579048 \\
0.5 0.96326421326193 \\
0.6 0.958075303867867 \\
0.7 0.952323758095315 \\
0.8 0.948275792720204 \\
0.9 0.945685258765125 \\
1 0.951934546119433 \\
};
\addlegendentry{\gls{ev}}
\addplot [thick, color3, mark=*, mark size=1, mark options={solid}]
table[row sep=\\] {
0 0.974866646287328 \\
0.1 0.971116839080278 \\
0.2 0.967067375440897 \\
0.3 0.961443101583703 \\
0.4 0.954235648212375 \\
0.5 0.947790024230668 \\
0.6 0.94107523802688 \\
0.7 0.935490996488859 \\
0.8 0.92854942277039 \\
0.9 0.921833063581162 \\
1 0.915632571535544 \\
};
\addlegendentry{\gls{av}}
\addplot [thick, color4, mark=*, mark size=1, mark options={solid}]
table[row sep=\\] {
0 0.987468454991712 \\
0.1 0.983874728146732 \\
0.2 0.979051785592302 \\
0.3 0.973698590079999 \\
0.4 0.96580555597538 \\
0.5 0.958123090667976 \\
0.6 0.947163650605321 \\
0.7 0.93332638177767 \\
0.8 0.916548444389559 \\
0.9 0.898412855665215 \\
1 0.886747995562417 \\
};
\addlegendentry{\gls{np}}
\addplot [thick, color5, mark=*, mark size=1, mark options={solid}]
table[row sep=\\] {
0 0.987832408397913 \\
0.1 0.984331466495525 \\
0.2 0.979155541229112 \\
0.3 0.974029180279361 \\
0.4 0.96612863931558 \\
0.5 0.957896726973807 \\
0.6 0.946346396824299 \\
0.7 0.930242353984798 \\
0.8 0.911967759543884 \\
0.9 0.891236749665748 \\
1 0.878504737197228 \\
};
\addlegendentry{\gls{rv}}
\addplot [thick, white!50.1960784313725!black, mark=*, mark size=1, mark options={solid}]
table[row sep=\\] {
0 0.847008522600071 \\
0.1 0.848284193204556 \\
0.2 0.849304835321012 \\
0.3 0.846456934226886 \\
0.4 0.851890401029518 \\
0.5 0.847116241870385 \\
0.6 0.850303685607628 \\
0.7 0.846694694664512 \\
0.8 0.84799407480579 \\
0.9 0.850062635259183 \\
1 0.848397516218121 \\
};
\addlegendentry{\gls{sa}}
\addplot [thick, color6, mark=*, mark size=1, mark options={solid}]
table[row sep=\\] {
0 0.828339454476499 \\
0.1 0.827042038154741 \\
0.2 0.822347454335365 \\
0.3 0.815681898366073 \\
0.4 0.80616942867992 \\
0.5 0.794835133567488 \\
0.6 0.788795774235565 \\
0.7 0.787875367690774 \\
0.8 0.783428139652128 \\
0.9 0.788177193957461 \\
1 0.800640278285634 \\
};
\addlegendentry{\gls{ml}}
\legend{}
\end{axis}

\end{tikzpicture}
}
	\subfloat[Decrease independence of the individual agents.\label{fig:beta}]{
\begin{tikzpicture}

\definecolor{color0}{rgb}{0.270588235294118,0.250980392156863,0.811764705882353}
\definecolor{color1}{rgb}{0.658823529411765,0.23921568627451,0.23921568627451}
\definecolor{color2}{rgb}{0.870588235294118,0.188235294117647,0.164705882352941}
\definecolor{color3}{rgb}{0.196078431372549,0.901960784313726,0.172549019607843}
\definecolor{color4}{rgb}{0.929411764705882,0.572549019607843,0.101960784313725}
\definecolor{color5}{rgb}{0.870588235294118,0.87843137254902,0.274509803921569}
\definecolor{color6}{rgb}{0.141176470588235,0.533333333333333,0.929411764705882}

\begin{axis}[
height=6cm, width=8cm,
legend cell align={left},
legend style={mylegend, fivecolumns,
  at={(0.97,0.03)},
  anchor=south east,
},
tick align=outside,
tick pos=left,
x grid style={white!69.0196078431373!black},
xlabel={\strut Absorption \(\displaystyle \beta\)},
x label style = {yshift=.15cm},
xmajorgrids,
xmin=0, xmax=1,
xtick style={color=black},
y grid style={white!69.0196078431373!black},
yticklabels={},
ymajorgrids,
ymin=0.7, ymax=1,
ytick style={draw=none}
]
\addplot [thick, white!27.0588235294118!black, mark=*, mark size=1, mark options={solid}]
table[row sep=\\] {
0 0.953234546628265 \\
0.1 0.946029195366857 \\
0.2 0.935811756930908 \\
0.3 0.924364834654783 \\
0.4 0.909050122493356 \\
0.5 0.892688704219556 \\
0.6 0.871768352736537 \\
0.7 0.862573398613031 \\
0.8 0.853224840214099 \\
0.9 0.852720714145091 \\
1 0.850520504364629 \\
};
\addlegendentry{\gls{ga}}
\addplot [thick, color0, mark=*, mark size=1, mark options={solid}]
table[row sep=\\] {
0 0.950327950359675 \\
0.1 0.944294010596841 \\
0.2 0.933491408946952 \\
0.3 0.922761070726215 \\
0.4 0.906965519431279 \\
0.5 0.891445648733804 \\
0.6 0.871486735354563 \\
0.7 0.862273484894629 \\
0.8 0.853987489701385 \\
0.9 0.853132013435142 \\
1 0.850516157781587 \\
};
\addlegendentry{\gls{ml+}}
\addplot [thick, color1, mark=*, mark size=1, mark options={solid}]
table[row sep=\\] {
0 0.952188919876275 \\
0.1 0.944572174282024 \\
0.2 0.93272022136057 \\
0.3 0.919704631215027 \\
0.4 0.901339005227755 \\
0.5 0.883948139603827 \\
0.6 0.865697675732371 \\
0.7 0.859009865633484 \\
0.8 0.852029501627037 \\
0.9 0.851232597432184 \\
1 0.850520504364629 \\
};
\addlegendentry{\gls{ev+}}
\addplot [thick, color2, mark=*, mark size=1, mark options={solid}]
table[row sep=\\] {
0 0.951934546119433 \\
0.1 0.944073871058007 \\
0.2 0.931688690957001 \\
0.3 0.918707475865677 \\
0.4 0.900407860220078 \\
0.5 0.883332490637319 \\
0.6 0.865205543663935 \\
0.7 0.858399734228747 \\
0.8 0.851175810671828 \\
0.9 0.851555289679133 \\
1 0.850697080670263 \\
};
\addlegendentry{\gls{ev}}
\addplot [thick, color3, mark=*, mark size=1, mark options={solid}]
table[row sep=\\] {
0 0.915632571535544 \\
0.1 0.908384208266981 \\
0.2 0.894294660106712 \\
0.3 0.884825785835027 \\
0.4 0.871810793011281 \\
0.5 0.8624456696851 \\
0.6 0.852960137860063 \\
0.7 0.851318419763286 \\
0.8 0.850023243054072 \\
0.9 0.851211194966746 \\
1 0.850516917722335 \\
};
\addlegendentry{\gls{av}}
\addplot [thick, color4, mark=*, mark size=1, mark options={solid}]
table[row sep=\\] {
0 0.886747995562417 \\
0.1 0.881446335565919 \\
0.2 0.874536809264869 \\
0.3 0.869326007372631 \\
0.4 0.863926009141228 \\
0.5 0.860254683880802 \\
0.6 0.851820019498971 \\
0.7 0.850959003174032 \\
0.8 0.849956954097714 \\
0.9 0.85119584834325 \\
1 0.850538417375062 \\
};
\addlegendentry{\gls{np}}
\addplot [thick, color5, mark=*, mark size=1, mark options={solid}]
table[row sep=\\] {
0 0.878504737197228 \\
0.1 0.872846417062763 \\
0.2 0.869314853228899 \\
0.3 0.866315953893231 \\
0.4 0.862745928064356 \\
0.5 0.859320161572236 \\
0.6 0.85152965988491 \\
0.7 0.850805665606296 \\
0.8 0.849898175908573 \\
0.9 0.851241318252352 \\
1 0.850529281478151 \\
};
\addlegendentry{\gls{rv}}
\addplot [thick, white!50.1960784313725!black, mark=*, mark size=1, mark options={solid}]
table[row sep=\\] {
0 0.848397516218121 \\
0.1 0.847207828462096 \\
0.2 0.84599275356688 \\
0.3 0.848703728367945 \\
0.4 0.849440763730124 \\
0.5 0.849866800907979 \\
0.6 0.846288371472185 \\
0.7 0.847108284000309 \\
0.8 0.847483450897246 \\
0.9 0.849997506134067 \\
1 0.848777614520099 \\
};
\addlegendentry{\gls{sa}}
\addplot [thick, color6, mark=*, mark size=1, mark options={solid}]
table[row sep=\\] {
0 0.800640278285634 \\
0.1 0.797426809668553 \\
0.2 0.798591442296427 \\
0.3 0.800491066852415 \\
0.4 0.797541051087973 \\
0.5 0.802401236596542 \\
0.6 0.806870640945753 \\
0.7 0.824610274581425 \\
0.8 0.83656703666971 \\
0.9 0.847959924219325 \\
1 0.848432672767057 \\
};
\addlegendentry{\gls{ml}}
\end{axis}

\end{tikzpicture}
}
	\caption{Modulating the correlations, from all agents being independent ($\alpha=0$ on Figure~\ref{fig:alpha}) to one unique correlated group ($\beta=1$ on Figure~\ref{fig:beta}). The values $\alpha=1$ on Figure~\ref{fig:alpha} and $\beta=0$ on Figure~\ref{fig:beta} correspond to the reference scenario (Figure~\ref{fig:base}). \label{fig:alphabeta}}
\end{figure*}

In Figure~\ref{fig:noise}, we change the intensities $\sigma_d$ and $\sigma_f$, corresponding to the distinct noise and the feature noise respectively, from 0.1 to 10, while the variance of the true utility distribution stays equal to~1. Unsurprisingly, when the noise increases, the performance degrades for all rules. \gls{ml+} stays on par with \gls{ga} in all settings. \gls{ev+} drops slightly only when both $\sigma_d$ and $\sigma_f$ are high. In addition to this case, the untrained version \gls{ev} is also sub-optimal when $\sigma_d = \sigma_f = 1$. However, it remains the best of all untrained rules for all settings.

\renewcommand{\thefigure}{5}
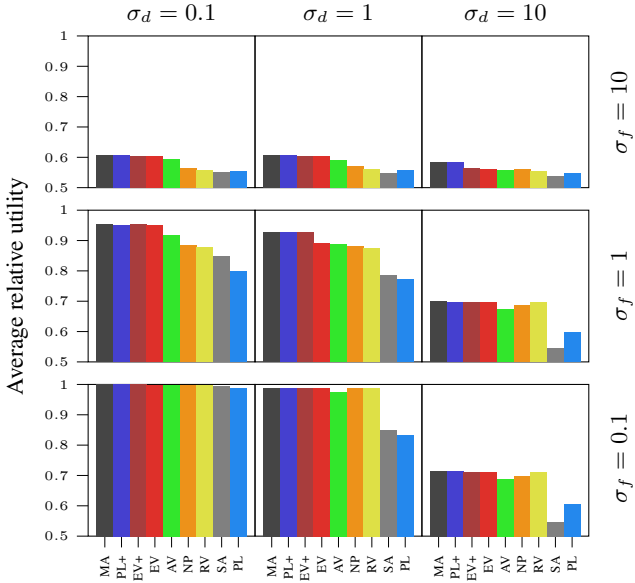
\begin{figure}[h]
\begin{tikzpicture}

\definecolor{color0}{rgb}{0.270588235294118,0.250980392156863,0.811764705882353}
\definecolor{color1}{rgb}{0.658823529411765,0.23921568627451,0.23921568627451}
\definecolor{color2}{rgb}{0.870588235294118,0.188235294117647,0.164705882352941}
\definecolor{color3}{rgb}{0.196078431372549,0.901960784313726,0.172549019607843}
\definecolor{color4}{rgb}{0.929411764705882,0.572549019607843,0.101960784313725}
\definecolor{color5}{rgb}{0.870588235294118,0.87843137254902,0.274509803921569}
\definecolor{color6}{rgb}{0.141176470588235,0.533333333333333,0.929411764705882}

\begin{groupplot}[
  group style={
    group size=3 by 3, 
    horizontal sep=0cm,
    vertical sep=.3cm,
  }, 
  width=3.8cm, 
  height=3.6cm
]
\node[rotate=90] at (7.1, 1) {$\sigma_f = 10$};
\node[rotate=90] at (7.1, -1.3) {$\sigma_f = 1$};
\node[rotate=90] at (7.1, -3.6) {$\sigma_f = 0.1$};

\nextgroupplot[
tick align=outside,
tick pos=left,
title={$\sigma_d = 0.1$},
title style={yshift=-1ex},
x grid style={white!69.0196078431373!black},
xmin=-1, xmax=9,
xtick style={draw=none},
xticklabels={},
xtick style={color=black},
y grid style={white!69.0196078431373!black},
ymin=0.5, ymax=1,
ytick style={color=black},
ytick distance=0.1,
yticklabel style={font=\tiny},
]
\draw[draw=none,fill=white!27.0588235294118!black] (axis cs:-0.5,0) rectangle (axis cs:0.5,0.607518318524233);
\draw[draw=none,fill=color0] (axis cs:0.5,0) rectangle (axis cs:1.5,0.607361691728188);
\draw[draw=none,fill=color1] (axis cs:1.5,0) rectangle (axis cs:2.5,0.604974698699211);
\draw[draw=none,fill=color2] (axis cs:2.5,0) rectangle (axis cs:3.5,0.602376818387246);
\draw[draw=none,fill=color3] (axis cs:3.5,0) rectangle (axis cs:4.5,0.592988781365341);
\draw[draw=none,fill=color4] (axis cs:4.5,0) rectangle (axis cs:5.5,0.56568322883284);
\draw[draw=none,fill=color5] (axis cs:5.5,0) rectangle (axis cs:6.5,0.556211016258003);
\draw[draw=none,fill=white!50.1960784313725!black] (axis cs:6.5,0) rectangle (axis cs:7.5,0.550127735615499);
\draw[draw=none,fill=color6] (axis cs:7.5,0) rectangle (axis cs:8.5,0.554415243522934);

\nextgroupplot[
tick align=outside,
tick pos=left,
title={$\sigma_d = 1$},
title style={yshift=-1ex},
x grid style={white!69.0196078431373!black},
xmin=-1, xmax=9,
xtick style={draw=none},
xticklabels={},
xtick style={color=black},
y grid style={white!69.0196078431373!black},
ymin=0.5, ymax=1,
ytick style={draw=none},
yticklabels={}
]
\draw[draw=none,fill=white!27.0588235294118!black] (axis cs:-0.5,0) rectangle (axis cs:0.5,0.606469689713875);
\draw[draw=none,fill=color0] (axis cs:0.5,0) rectangle (axis cs:1.5,0.60622241710554);
\draw[draw=none,fill=color1] (axis cs:1.5,0) rectangle (axis cs:2.5,0.60356138722548);
\draw[draw=none,fill=color2] (axis cs:2.5,0) rectangle (axis cs:3.5,0.6022385222868);
\draw[draw=none,fill=color3] (axis cs:3.5,0) rectangle (axis cs:4.5,0.590173690979586);
\draw[draw=none,fill=color4] (axis cs:4.5,0) rectangle (axis cs:5.5,0.569245429979548);
\draw[draw=none,fill=color5] (axis cs:5.5,0) rectangle (axis cs:6.5,0.559451330384996);
\draw[draw=none,fill=white!50.1960784313725!black] (axis cs:6.5,0) rectangle (axis cs:7.5,0.548626695203655);
\draw[draw=none,fill=color6] (axis cs:7.5,0) rectangle (axis cs:8.5,0.556383635819707);

\nextgroupplot[
tick align=outside,
tick pos=left,
title={$\sigma_d = 10$},
title style={yshift=-1ex},
x grid style={white!69.0196078431373!black},
xmin=-1, xmax=9,
xtick style={draw=none},
xticklabels={},
xtick style={color=black},
y grid style={white!69.0196078431373!black},
ymin=0.5, ymax=1,
ytick style={draw=none},
yticklabels={}
]
\draw[draw=none,fill=white!27.0588235294118!black] (axis cs:-0.5,0) rectangle (axis cs:0.5,0.585527396632529);
\draw[draw=none,fill=color0] (axis cs:0.5,0) rectangle (axis cs:1.5,0.584867664622512);
\draw[draw=none,fill=color1] (axis cs:1.5,0) rectangle (axis cs:2.5,0.56536385482316);
\draw[draw=none,fill=color2] (axis cs:2.5,0) rectangle (axis cs:3.5,0.560040243351578);
\draw[draw=none,fill=color3] (axis cs:3.5,0) rectangle (axis cs:4.5,0.556690507438216);
\draw[draw=none,fill=color4] (axis cs:4.5,0) rectangle (axis cs:5.5,0.560610229452782);
\draw[draw=none,fill=color5] (axis cs:5.5,0) rectangle (axis cs:6.5,0.553811889871386);
\draw[draw=none,fill=white!50.1960784313725!black] (axis cs:6.5,0) rectangle (axis cs:7.5,0.537136854074378);
\draw[draw=none,fill=color6] (axis cs:7.5,0) rectangle (axis cs:8.5,0.547170952095678);

\nextgroupplot[
tick align=outside,
tick pos=left,
x grid style={white!69.0196078431373!black},
xmin=-1, xmax=9,
xtick style={draw=none},
xticklabels={},
xtick style={color=black},
y grid style={white!69.0196078431373!black},
ymin=0.5, ymax=1,
ytick style={color=black},
ytick distance=0.1,
yticklabel style={font=\tiny},
ylabel={Average relative utility},
]
\draw[draw=none,fill=white!27.0588235294118!black] (axis cs:-0.5,0) rectangle (axis cs:0.5,0.953199014376164);
\draw[draw=none,fill=color0] (axis cs:0.5,0) rectangle (axis cs:1.5,0.951831247052239);
\draw[draw=none,fill=color1] (axis cs:1.5,0) rectangle (axis cs:2.5,0.952785710216859);
\draw[draw=none,fill=color2] (axis cs:2.5,0) rectangle (axis cs:3.5,0.95209643627863);
\draw[draw=none,fill=color3] (axis cs:3.5,0) rectangle (axis cs:4.5,0.917148259384599);
\draw[draw=none,fill=color4] (axis cs:4.5,0) rectangle (axis cs:5.5,0.885860185364536);
\draw[draw=none,fill=color5] (axis cs:5.5,0) rectangle (axis cs:6.5,0.877198253256187);
\draw[draw=none,fill=white!50.1960784313725!black] (axis cs:6.5,0) rectangle (axis cs:7.5,0.848111899237714);
\draw[draw=none,fill=color6] (axis cs:7.5,0) rectangle (axis cs:8.5,0.799799404799069);

\nextgroupplot[
tick align=outside,
tick pos=left,
x grid style={white!69.0196078431373!black},
xmin=-1, xmax=9,
xtick style={draw=none},
xticklabels={},
xtick style={color=black},
y grid style={white!69.0196078431373!black},
ymin=0.5, ymax=1,
ytick style={draw=none},
yticklabels={}
]
\draw[draw=none,fill=white!27.0588235294118!black] (axis cs:-0.5,0) rectangle (axis cs:0.5,0.929392914286069);
\draw[draw=none,fill=color0] (axis cs:0.5,0) rectangle (axis cs:1.5,0.927016237389561);
\draw[draw=none,fill=color1] (axis cs:1.5,0) rectangle (axis cs:2.5,0.9279486945401);
\draw[draw=none,fill=color2] (axis cs:2.5,0) rectangle (axis cs:3.5,0.890306950334482);
\draw[draw=none,fill=color3] (axis cs:3.5,0) rectangle (axis cs:4.5,0.887474395735856);
\draw[draw=none,fill=color4] (axis cs:4.5,0) rectangle (axis cs:5.5,0.88256850841251);
\draw[draw=none,fill=color5] (axis cs:5.5,0) rectangle (axis cs:6.5,0.87428714105345);
\draw[draw=none,fill=white!50.1960784313725!black] (axis cs:6.5,0) rectangle (axis cs:7.5,0.786103401768549);
\draw[draw=none,fill=color6] (axis cs:7.5,0) rectangle (axis cs:8.5,0.772900664728193);

\nextgroupplot[
tick align=outside,
tick pos=left,
x grid style={white!69.0196078431373!black},
xmin=-1, xmax=9,
xtick style={draw=none},
xticklabels={},
xtick style={color=black},
y grid style={white!69.0196078431373!black},
ymin=0.5, ymax=1,
ytick style={draw=none},
yticklabels={}
]
\draw[draw=none,fill=white!27.0588235294118!black] (axis cs:-0.5,0) rectangle (axis cs:0.5,0.699126385393889);
\draw[draw=none,fill=color0] (axis cs:0.5,0) rectangle (axis cs:1.5,0.697859500087086);
\draw[draw=none,fill=color1] (axis cs:1.5,0) rectangle (axis cs:2.5,0.697617432783407);
\draw[draw=none,fill=color2] (axis cs:2.5,0) rectangle (axis cs:3.5,0.695795493551699);
\draw[draw=none,fill=color3] (axis cs:3.5,0) rectangle (axis cs:4.5,0.675306353307203);
\draw[draw=none,fill=color4] (axis cs:4.5,0) rectangle (axis cs:5.5,0.687050609968947);
\draw[draw=none,fill=color5] (axis cs:5.5,0) rectangle (axis cs:6.5,0.698329322774363);
\draw[draw=none,fill=white!50.1960784313725!black] (axis cs:6.5,0) rectangle (axis cs:7.5,0.546693084913795);
\draw[draw=none,fill=color6] (axis cs:7.5,0) rectangle (axis cs:8.5,0.598828323750406);

\nextgroupplot[
tick align=outside,
tick pos=left,
x grid style={white!69.0196078431373!black},
xmin=-1, xmax=9,
xtick style={color=black},
xtick={0,1,2,3,4,5,6,7,8},
xticklabel style={rotate=90.0, font=\tiny},
xticklabels={\gls{ga},\gls{ml+},\gls{ev+},\gls{ev},\gls{av},\gls{np},\gls{rv},\gls{sa},\gls{ml}},
y grid style={white!69.0196078431373!black},
ymin=0.5, ymax=1,
ytick style={color=black},
ytick distance=0.1,
yticklabel style={font=\tiny},
]
\draw[draw=none,fill=white!27.0588235294118!black] (axis cs:-0.5,0) rectangle (axis cs:0.5,0.998990962426529);
\draw[draw=none,fill=color0] (axis cs:0.5,0) rectangle (axis cs:1.5,0.999008226325576);
\draw[draw=none,fill=color1] (axis cs:1.5,0) rectangle (axis cs:2.5,0.998859365379886);
\draw[draw=none,fill=color2] (axis cs:2.5,0) rectangle (axis cs:3.5,0.998493782130373);
\draw[draw=none,fill=color3] (axis cs:3.5,0) rectangle (axis cs:4.5,0.997965140705221);
\draw[draw=none,fill=color4] (axis cs:4.5,0) rectangle (axis cs:5.5,0.997963250091976);
\draw[draw=none,fill=color5] (axis cs:5.5,0) rectangle (axis cs:6.5,0.997963250091976);
\draw[draw=none,fill=white!50.1960784313725!black] (axis cs:6.5,0) rectangle (axis cs:7.5,0.994444655447443);
\draw[draw=none,fill=color6] (axis cs:7.5,0) rectangle (axis cs:8.5,0.98839826312717);

\nextgroupplot[
tick align=outside,
tick pos=left,
x grid style={white!69.0196078431373!black},
xmin=-1, xmax=9,
xtick style={color=black},
xtick={0,1,2,3,4,5,6,7,8},
xticklabel style={rotate=90.0, font=\tiny},
xticklabels={\gls{ga},\gls{ml+},\gls{ev+},\gls{ev},\gls{av},\gls{np},\gls{rv},\gls{sa},\gls{ml}},
y grid style={white!69.0196078431373!black},
ymin=0.5, ymax=1,
ytick style={draw=none},
yticklabels={}
]
\draw[draw=none,fill=white!27.0588235294118!black] (axis cs:-0.5,0) rectangle (axis cs:0.5,0.987262051428852);
\draw[draw=none,fill=color0] (axis cs:0.5,0) rectangle (axis cs:1.5,0.985892458015029);
\draw[draw=none,fill=color1] (axis cs:1.5,0) rectangle (axis cs:2.5,0.985677127873595);
\draw[draw=none,fill=color2] (axis cs:2.5,0) rectangle (axis cs:3.5,0.986358966045011);
\draw[draw=none,fill=color3] (axis cs:3.5,0) rectangle (axis cs:4.5,0.974103119732411);
\draw[draw=none,fill=color4] (axis cs:4.5,0) rectangle (axis cs:5.5,0.986772017873613);
\draw[draw=none,fill=color5] (axis cs:5.5,0) rectangle (axis cs:6.5,0.98708624106152);
\draw[draw=none,fill=white!50.1960784313725!black] (axis cs:6.5,0) rectangle (axis cs:7.5,0.848006446506946);
\draw[draw=none,fill=color6] (axis cs:7.5,0) rectangle (axis cs:8.5,0.831616535679557);

\nextgroupplot[
tick align=outside,
tick pos=left,
x grid style={white!69.0196078431373!black},
xmin=-1, xmax=9,
xtick style={color=black},
xtick={0,1,2,3,4,5,6,7,8},
xticklabel style={rotate=90.0, font=\tiny},
xticklabels={\gls{ga},\gls{ml+},\gls{ev+},\gls{ev},\gls{av},\gls{np},\gls{rv},\gls{sa},\gls{ml}},
y grid style={white!69.0196078431373!black},
ymin=0.5, ymax=1,
ytick style={draw=none},
yticklabels={}
]
\draw[draw=none,fill=white!27.0588235294118!black] (axis cs:-0.5,0) rectangle (axis cs:0.5,0.712607947281673);
\draw[draw=none,fill=color0] (axis cs:0.5,0) rectangle (axis cs:1.5,0.712255088085978);
\draw[draw=none,fill=color1] (axis cs:1.5,0) rectangle (axis cs:2.5,0.709257073765654);
\draw[draw=none,fill=color2] (axis cs:2.5,0) rectangle (axis cs:3.5,0.708636851362344);
\draw[draw=none,fill=color3] (axis cs:3.5,0) rectangle (axis cs:4.5,0.687673600957852);
\draw[draw=none,fill=color4] (axis cs:4.5,0) rectangle (axis cs:5.5,0.698266257063247);
\draw[draw=none,fill=color5] (axis cs:5.5,0) rectangle (axis cs:6.5,0.709882535035109);
\draw[draw=none,fill=white!50.1960784313725!black] (axis cs:6.5,0) rectangle (axis cs:7.5,0.546063651641049);
\draw[draw=none,fill=color6] (axis cs:7.5,0) rectangle (axis cs:8.5,0.606037973243906);
\end{groupplot}

\end{tikzpicture}

	\caption{Changing the noise intensity\label{fig:noise}}
\end{figure}

\vspace{-.1cm}
\subsection{Soft Partition of the Agents}\label{sec:correlated-groups}
\vspace{-.1cm}

The reference scenario represents a clear partition of the agents into a group of $20$ and $4$ individuals. Here we modify the embedding $E$ to account for more subtle correlations.

For example, the correlation between two agents $a_{i}$ and $a_{i'}$ from the group may depend on $i$ and $i'$ (e.g. the agents are algorithms and their indices represent a varying hyperparameter). For some $\alpha \in [0, 1]$, this can be modeled by a $\medmuskip=0mu 24 \times 24$ embedding matrix defined by blocks as $E_\alpha = \left( \begin{array}{cc} G_\alpha & 0 \\ 0 & I_4 \end{array} \right)$, 
where $(G_\alpha)_{i,i'} \propto \alpha^{|i-i'|}$.
In this model, the correlation between two agents of the group depends on the distance between their respective indices. The model parameter $\alpha$, called \emph{cohesion}, controls the group correlations: if $\alpha=0$, all agents of the group are independent; if $\alpha=1$, the model is equivalent to the reference scenario.

Another case that makes the partition less binary is if the independent agents can be influenced, to some degree, by the opinion of the group.
For some  $\beta \in [0, 1]$, this can be modeled by a $24 \times 5$ embedding matrix 
defined by blocks as  $E_\beta = \left( \begin{array}{cc} \mathds{1}_{20\times 1} & 0 \\ c\beta \mathds{1}_{4\times 1} & c(1-\beta)I_4 \end{array} \right)$, with $c=\frac{1}{\sqrt{\beta^2 + (1-\beta)^2}}$.
The model parameter $\beta$, which we call \emph{absorption}, controls the influence of the group on the independent agents: if $\beta=0$, which corresponds to the reference scenario, the influence is null; if $\beta=1$, all $24$ agents form a unique group.

Figure~\ref{fig:alphabeta} show the evolution of the utility when $\alpha$ and $\beta$ vary respectively. It can be read as one unique figure ranging from $24$ independent agents ($\alpha=0$) to one unique group ($\beta=1$), with the reference scenario in the middle ($\alpha=1$ or $\beta=0$). The results are in line with what we have observed so far: if the agents are fully independent or correlated, any decent rule like \gls{rv} is optimal, but in the presence of non-trivial correlations, the rules \gls{ml+} and \gls{ev+} are virtually indistinguishable from \gls{ga}. \gls{ev} has slightly lower performance (the difference is of the order of the margin of error) but clearly remains the best of all untrained rules. This remains true even if the correlations represent something more complex than a simple partition of the agents.

\vspace{-.2cm}
\section{Conclusion}
\label{sec:conclusion}
\vspace{-.1cm}

In the context of aggregating correlated noisy agents in a choice problem, we have proposed \acrlong{ev}, a method that embeds the agents according to the scores they produce, and we have compared its performance with a variety of other methods. Our main findings are that if a sufficient score history of the agents is available, then a maximum-likelihood approach is the best option. On the other hand, if the training set is limited, EV should be preferred, as it is robust and outperforms other untrained methods. 

\newpage
\section*{Acknowledgments}

This work was done at LINCS (\url{https://www.lincs.fr/}).

\noindent Théo Delemazure was supported by the PRAIRIE 3IA Institute under grant ANR-19-P3IA-0001 (e).

\end{document}